\documentclass{article}

\usepackage{PRIMEarxiv}
\usepackage{amsmath}
\usepackage{authblk}
\usepackage[utf8]{inputenc} 
\usepackage[T1]{fontenc}    
\usepackage{hyperref}       
\usepackage{url}            
\usepackage{booktabs}       
\usepackage{amsfonts}       
\usepackage{nicefrac}       
\usepackage{microtype}      
\usepackage{lipsum}
\usepackage{fancyhdr}       
\usepackage{graphicx}       
\graphicspath{{media/}}     
\usepackage{multirow}

\title{Sinoledge: A Knowledge Engine based on Logical Reasoning and Distributed Micro Services 
\thanks{\textit{\underline{Corresponding author}}: 
\textbf{Keke Tang \  email: \ tkk2012@gmail.com}} 
}

\author[\space\space 1,3]{Yining Huang \thanks{huangyining1987@gmail.com}}
\author[\space\space1]{Shaoze Lin \thanks{443452573@qq.com}}
\author[\space\space1]{Yijun Wei \thanks{eiiun@foxmail.com}}
\author[\space\space1,2]{*Keke Tang  \thanks{tkk2012@gmail.com}}
\affil[1]{Sinohealth research institution}
\affil[2]{Shenyang institute of computing technology, Chinese academy of sciences}
\affil[3]{School of Politics and Public Administration, South China Normal University}

\begin{document}
\maketitle

\begin{abstract}
We propose a knowledge engine called Sinoledge mainly for doctors, physicians, and researchers in medical field to organize thoughts, manage reasoning process, test and deploy to production environments effortlessly. Our proposal can be related to rule engine usually used in business or medical fields. More importantly, our proposal provides a user-friendly interface, a easy-maintain way of organizing knowledge, an understandable testing functionality and a highly available and efficient back-end architecture.
\end{abstract}


\section*{Introduction}
\label{sec:headings}
In recent years, medical resources in China have been in a state of short supply. Doctors from the top hospitals in modern cities have to face tedious consultation or surgical work every day. Meanwhile, some doctors also need to do scientific research related work for the development of medicine. However, in doctors’ daily work, a large part of the work is cumbersome but easy to use IT systems to improve efficiency, such as the management of medical terms, the collection of medical knowledge, and so on. The organization of medical knowledge is mostly for daily diagnosis and treatment in accordance with certain gold standards or guidelines, and many of them are based on some established rules to make inference. Therefore, it has become a hard demand to help doctors free themselves from tedious medical knowledge updating and sorting, knowledge reasoning calculations, and to be more concentrated in creative work in the diagnosis and treatment of intractable diseases or scientific research. Helping doctors to update and organize medical knowledge in a timely manner, including but not limited to the following scenario examples: 1. Automatic layout of follow-up plans. In the daily diagnosis and treatment process, doctors may encounter some special cases that require customized follow-up plans, but these plans are formulated in accordance with certain rules; 2. The updating of some clinical guidelines take much time from doctors to read, understand and apply, but the content of the clinical guidelines itself is easy to use established knowledge or reasoning rules to express; 3. In the follow-up process, for collecting data in medical researches as well as daily treatments, doctors have to design patient-friendly questions in order to make them easier to be understood. However, after getting answers from patients, they need to transform them into specific data or statistical values which it’s well-designed for work. In general, these transforming methods are based on fixed rules of statistics. The above are just some examples, there are many actual scenarios. On the other hand, since the traditional business system in the past needs to code the core business logic before putting it into use, this method lacks a certain degree of flexibility, and the established logic is hard to maintain and cannot be applied to flexible business scenarios. Therefore, doctors' custom inference rules, business logic flow, and seamless connection with business systems are another core requirement.
Based on the requirements mentioned above, we design a knowledge engine. It includes an intuitive, flexible, easy-to-understand, and highly maneuverable knowledge input interface, testing module, powerful knowledge reasoning services based on descriptive logic reasoning, and a high available, auto-scaled back-end architecture. The knowledge engine can help doctors intuitively collect, organize and maintain daily medical knowledge, facilitate reusing and avoid duplication of work, and can seamlessly connect to business systems by using non-coding methods. Using a knowledge engine can improve the work efficiency of doctors, avoid dealing with complex and tedious business, and make them more focused on diagnosis, treatment and scientific research. And the threshold for using the knowledge engine is low. It only requires doctors to have a basic understanding of general logic calculations, and then get started through quick training. In addition, thanks to the flexibility of the knowledge engine, more needs from business scenarios can also be met.

\section*{Related work}
\label{sec:headings}
A rule-based engine for reasoning and calculating is widely used in medical field because of its user-friendly interface and maintenance. Early in 2006, \cite{RN32} found that the execution process in business could be introduced in medical field because of the features and advantages. Specifically, they use BPEL web service language to describe decision flows in CDS rule engine. In 2007, \cite{RN25} present an approach to align with the AMIA roadmap of optimizing CDS interventions by separating the rules and the execution code. Moreover, the rules are in a standard formation and customizable which leads to simpler maintenance.

Most of the use cases of rule-based engine in medical field are focus on some specific aspects of CDS. In the study of \cite{RN29}, CDSS rule engine is responsible for generating and transferring information to patients' mobile phone in order to give individualized medical instructions. \cite{RN14} use rule engine for reasoning and ontologies for encapsulating medical knowledge in the task of drug-drug and drug-disease interaction. \cite{RN5} represents a use case of adverse drug events detection based on rule engine. \cite{RN10} use Drools rule engine to support clinical decision of chronic kidney disease based on EHR data. In addition, experiments for measuring performance differences are conducted and presented in this paper. In \cite{RN24}, they demonstrate a health care management system based on rule engine enabled CDS system, which is helpful in archiving glycemic control for patients with diabetes. Furthermore, a randomized controlled trial is conducted for proving that system is effective to management for older patients with type 2 diabetes. \cite{RN11} implement a medical rule engine, by using it, they model a clinical guideline for the prevention of mother-to-child transmission of hepatitis B as a clinical decision flow. Later on, they improve their system by separating business logic from medical knowledge, because the clinical workflows can be reused and maintained more easily \cite{RN22}. \cite{RN18} propose a semantic web framework for infection level recommendation and risk assessment. \cite{RN19} present a study for identifying high-risk PICU patients by using rule engine based CDS system and it archives good performance on correctness. 

Some other work of CDS focus on technical aspect. \cite{RN13} validate the clinical knowledge and execution performance of two knowledge engines and it turns out that the one called uBrain is more effective. \cite{RN26} present development environment, compiler, rule engine and application server for a CDS system based on a widely used clinical and scientific standard called Arden Syntax. This paper also indicate the problem of interaction between CDS system and health information systems. To address this problem, they use HL7 standard GELLO which provides interface and query language to the information system. \cite{RN33} presents an application based on rule engine for medical treatment procedures. Mainly, data processing can be divided into stages of transforming from personal medical devices to standard form and executing for various support actions. A example use case of cardiovascular diseases is shown in it. In the study of \cite{RN24}, during the follow-up process, rule engine in the system is responsible for generating and transferring structured information. 

\cite{RN7} design a system for CHF patients health status monitoring and follow-up based on rule engine. \cite{RN21} implement a data structure model called vMR to meet the requirement of CDS rule engine. Their work could be seen as adaptation layer between patient records and CDS systems. In \cite{RN9} and \cite{RN2}, use rule engine for managing business logic of the mapping service which is responsible for linking information structures of different standards and technologies. \cite{RN1} present a scalable architecture for rule engine based CDS system.

Apart from CDS, rule engine can also be applied in other aspects. \cite{RN8} propose an adjustable rule engine for filtering and generating new multilingual UMLS terms. \cite{RN27} show a use case of forming transportation plans, including dispatching available ambulances and helicopters and even taking special equipments in different situations when incidents happened with a customizable rule engine. \cite{RN6} use rule engine with decision tables to be the core of an organ allocation system. this paper indicates that the rule engine is easier to implement and successful in verifying the correctness, completeness and consistency. \cite{RN3} evaluate the performance of rule engine called iLog by decoupling it into there major parts, which are rule engine, application and support logic. Furthermore, they realize that the bottle neck is delivering patient data to rule engine for execution.  

\section*{Features}
\label{sec:headings}
Knowledge engine is a micro service-based and container-based knowledge inputting and logical reasoning engine. In the following chapter, some details are going to be introduced and discussed, including the functionality, system architecture, principle of logical computation, computational modules, data model, solution of storing data  and design of availability. 

\subsection*{Functionality}
The entire system of knowledge engine can be divided into different modules and sub-modules. Each of them is responsible for different procedures. 

\subsubsection*{Management of Calculation Pattern}
In knowledge engine, logical reasoning is the major inference methods. But it’s not adequate to adapt to outer business systems merely with logical reasoning. Hence, basic calculation, including numerical calculation and set operations need to be included in it. In calculation pattern management module, user can create calculation patterns by filling with necessary information and some specific types of variables according to the requirement brought by the selected function. For example, if we want to transforming a blood pressure value into a specific concept “high blood pressure” when its value is higher than a certain threshold (e.g. 140) , firstly, we need to select a “greater than” function and specify the numerical variable “blood pressure” and the threshold 140 for comparing; secondly, we need to select the output concept of “high blood pressure”. The requirement of input and output parameters are decided by the function chosen. Moreover, the calculation process will be shown in the preview area intuitively. After all those are set, the defined calculation pattern can be used in organizing the entire decision flow. Besides, the functions selected are built in the system. Usually, they are some common and frequently used functions, such as “greater than”, “less than” etc. We are also considering some more user-friendly ways of designing complex calculations functions.

\begin{figure}[htp]
    \centering
    \includegraphics[width=12cm]{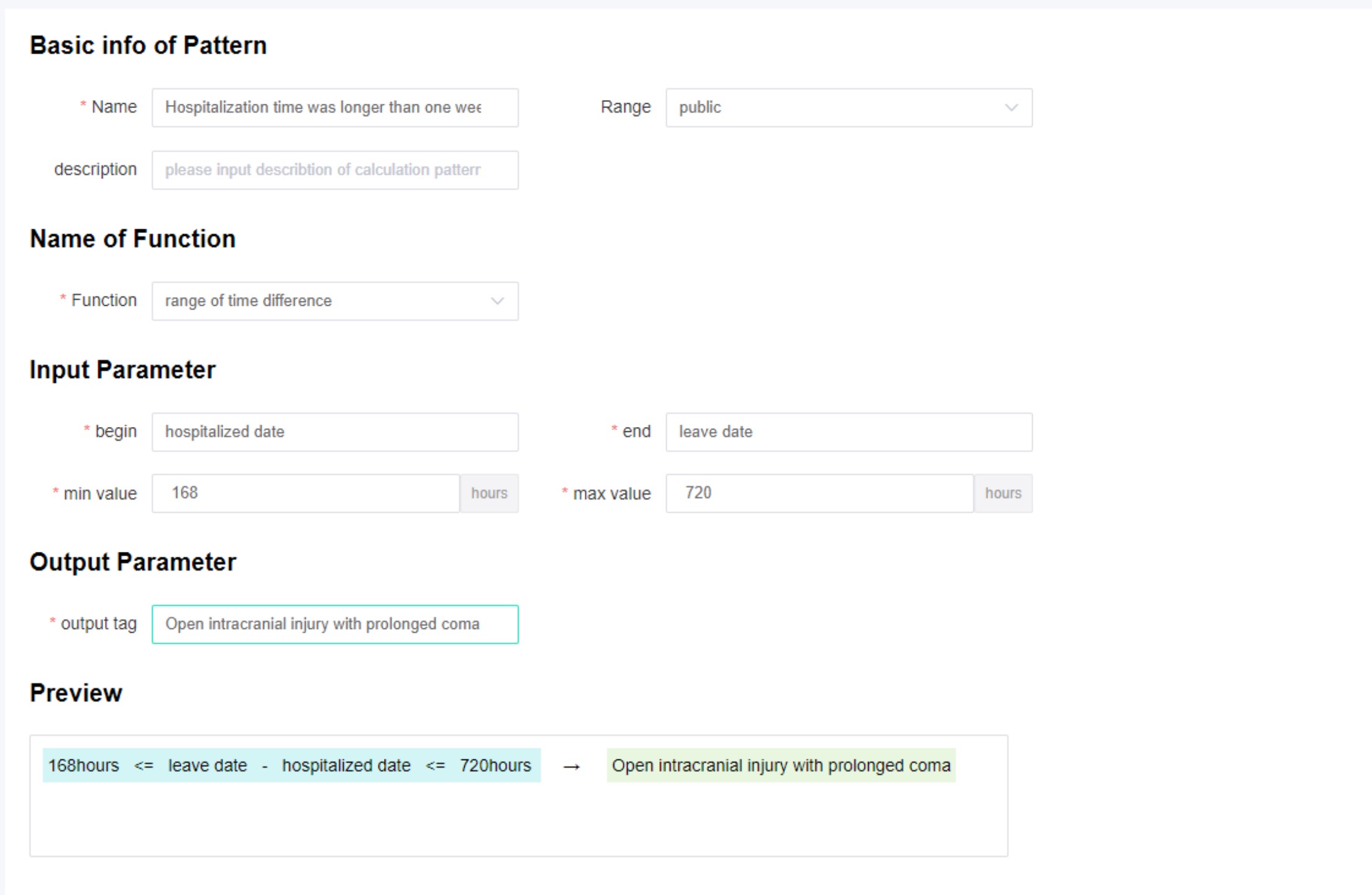}
    \caption{Setting Calculation Pattern}
    \label{fig:galaxy}
\end{figure}

\subsubsection*{Management of Decision Flow}
One of the core functions of knowledge engine is the management of decision flow. Decision flow is a way of organizing decision blocks. Decision blocks are ordered in a decision flow. Splitting a decision flow into small blocks makes it more understandable at a higher level and blocks can be reused more easily for different tasks. The management function of decision flow includes building, testing and version management.

\subsubsection*{Building}
We can build the main flow of ordered decisions by giving detail information and the order among them. After it is finished, more detail of logical reasoning process can be filled in different decision blocks. In a particular decision block, user can express the reasoning process in the form of simple descriptive logic with the basic logical operator. Moreover, complex calculation, including numerical and set operation, can be expressed in the form of calculation pattern which is mentioned in the last sub chapter.

\begin{figure}[htp]
    \centering
    \includegraphics[height=8cm]{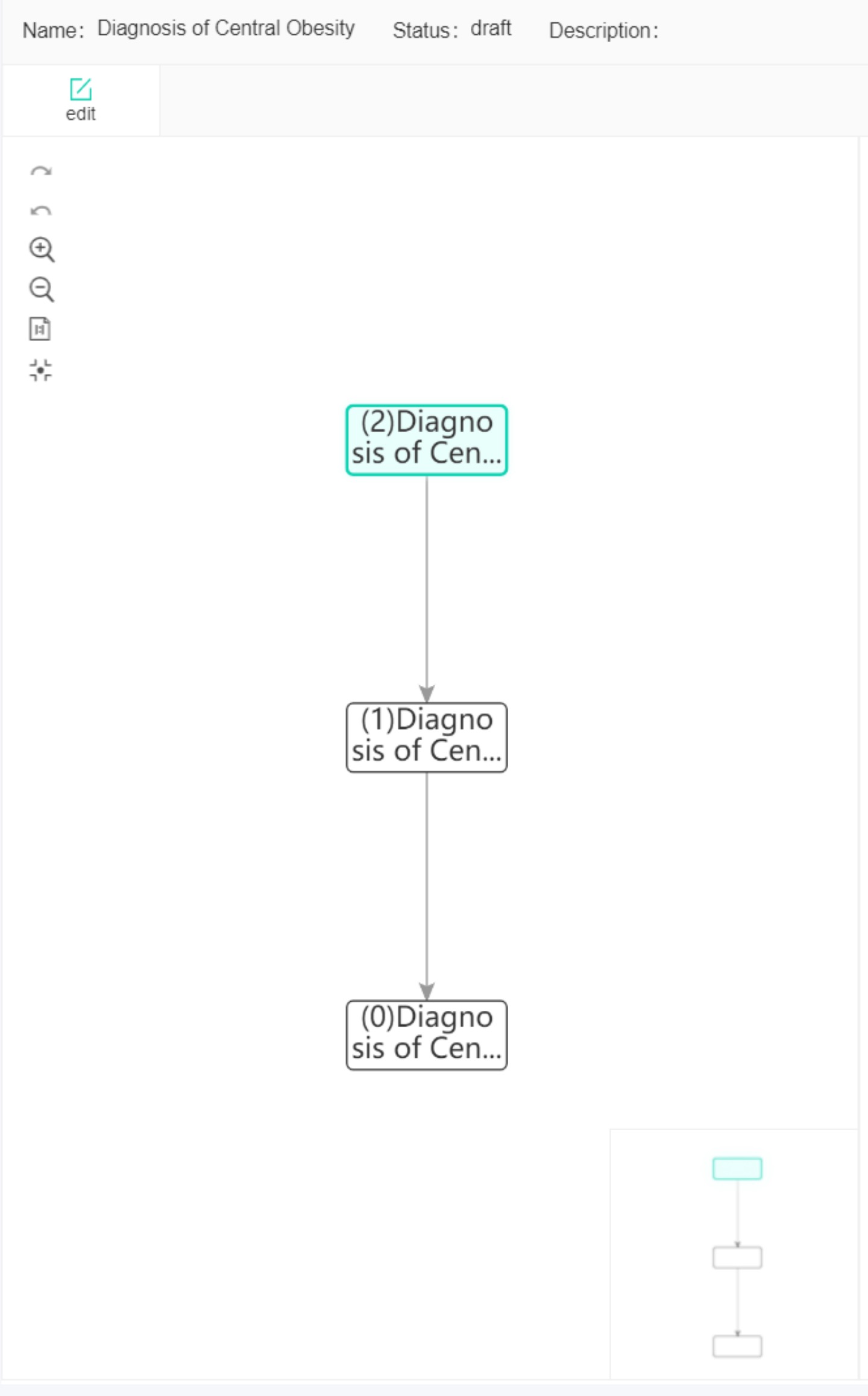}
    \caption{Decision Flow}
    \label{fig:galaxy}
\end{figure}

\begin{figure}[htp]
    \centering
    \includegraphics[width=12cm]{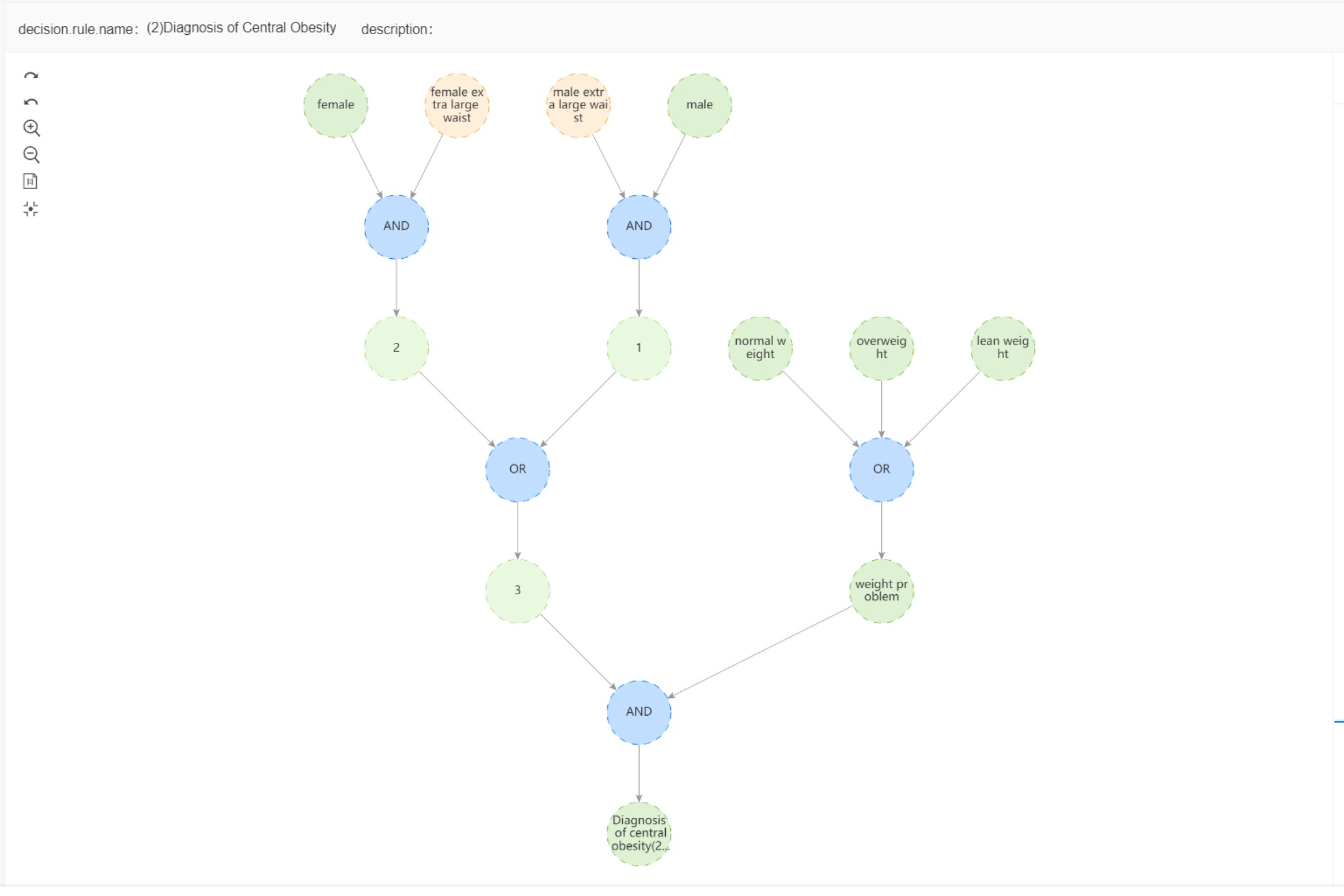}
    \caption{Logical Reasoning Process}
    \label{fig:galaxy}
\end{figure}

\begin{figure}[htp]
    \centering
    \includegraphics[width=10cm]{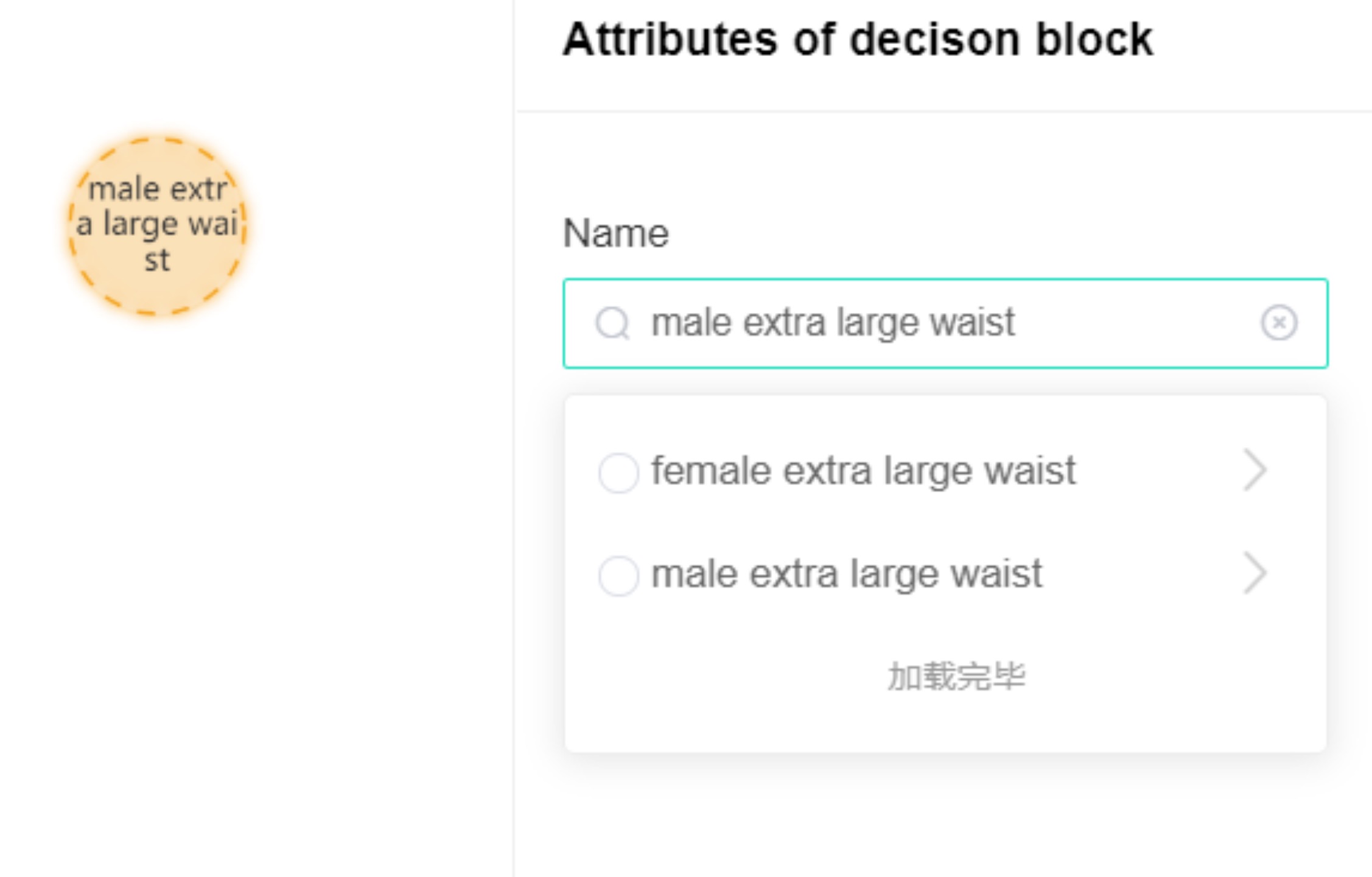}
    \caption{Setting existing calculation pattern}
    \label{fig:galaxy}
\end{figure}

\subsubsection*{Testing}
When the decision flow and its major decision blocks are built, testing for measuring the mismatch between actual calculation result and the expectation need to be done next. There are two different testing process. One is called online test. It’s for quick validating the reasoning logic. By using the template generated from the content of the block, users can validate their work. More importantly, the pathway of calculation and main reasons cause the mismatch result will be shown in order to let users recognize them as fast as possible. 

\begin{figure}[htp]
    \centering
    \includegraphics[width=12cm]{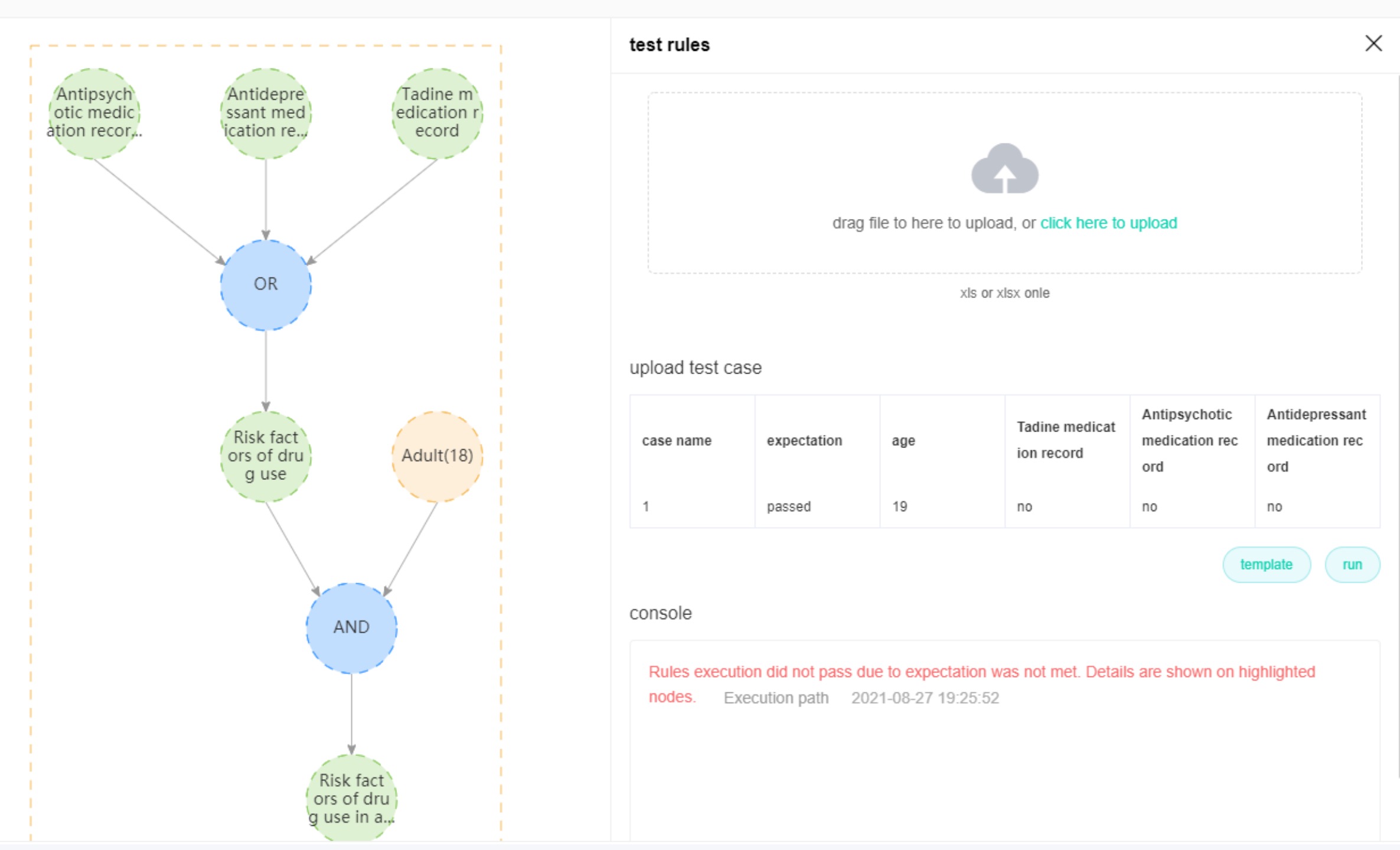}
    \caption{Online Test}
    \label{fig:galaxy}
\end{figure}

Another one is called test in batch. Literally, user can perform testing with a batch of data. It can also be performed in the way of using template. In which, multiple cases testing can be performed, and inputs in every case can be shown. The function of tracking the reasoning path is included as well. Unlike online test, testing in batch is designed for performing test activities with a large amount of data. Hence, data can be managed in the system for testing many times. Nevertheless, statistics for testing records, time consumed, status, accuracy, will be included for showing the overall situation. 

\begin{figure}[htp]
    \centering
    \includegraphics[width=12cm]{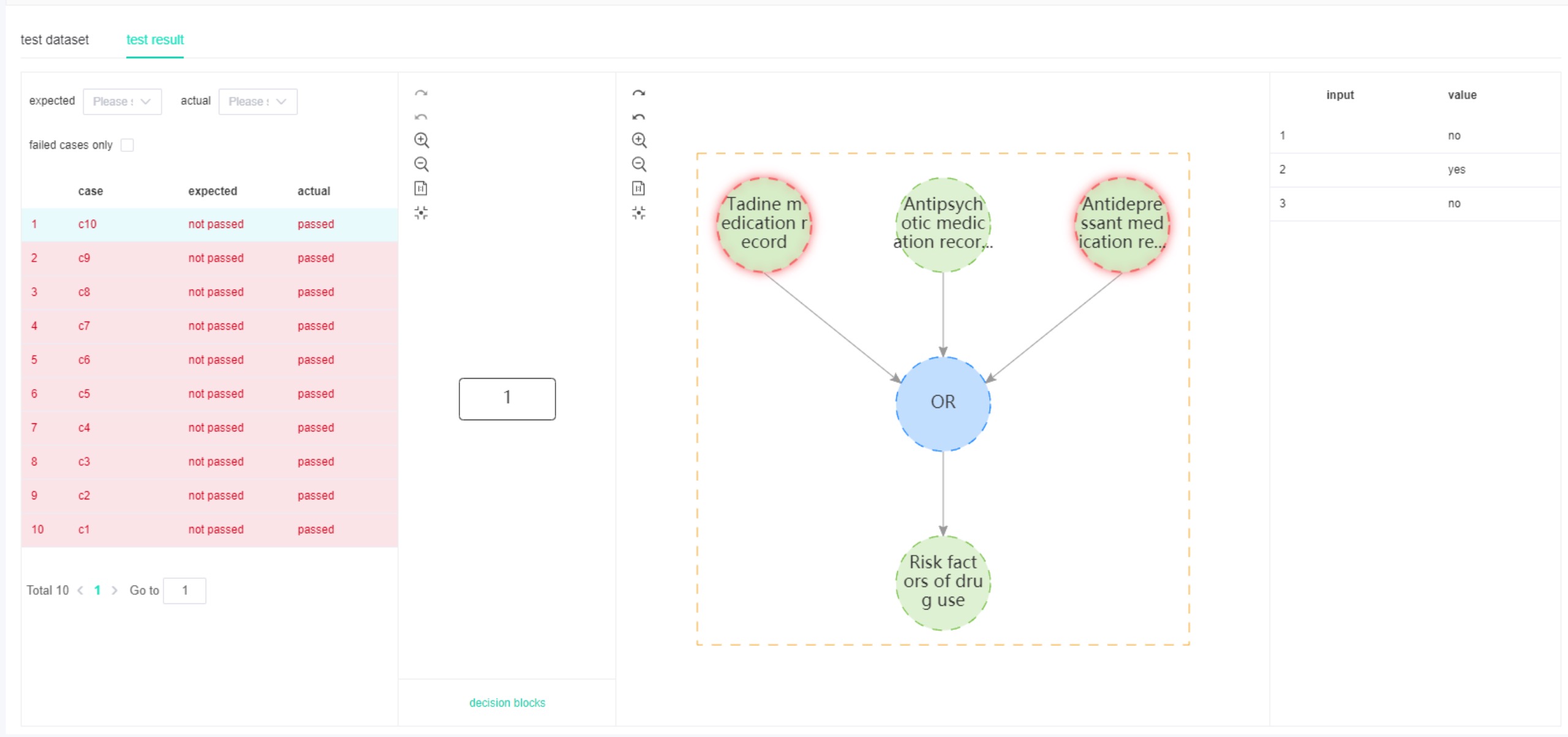}
    \caption{Test In Batch}
    \label{fig:galaxy}
\end{figure}

\begin{figure}[htp]
    \centering
    \includegraphics[width=12cm]{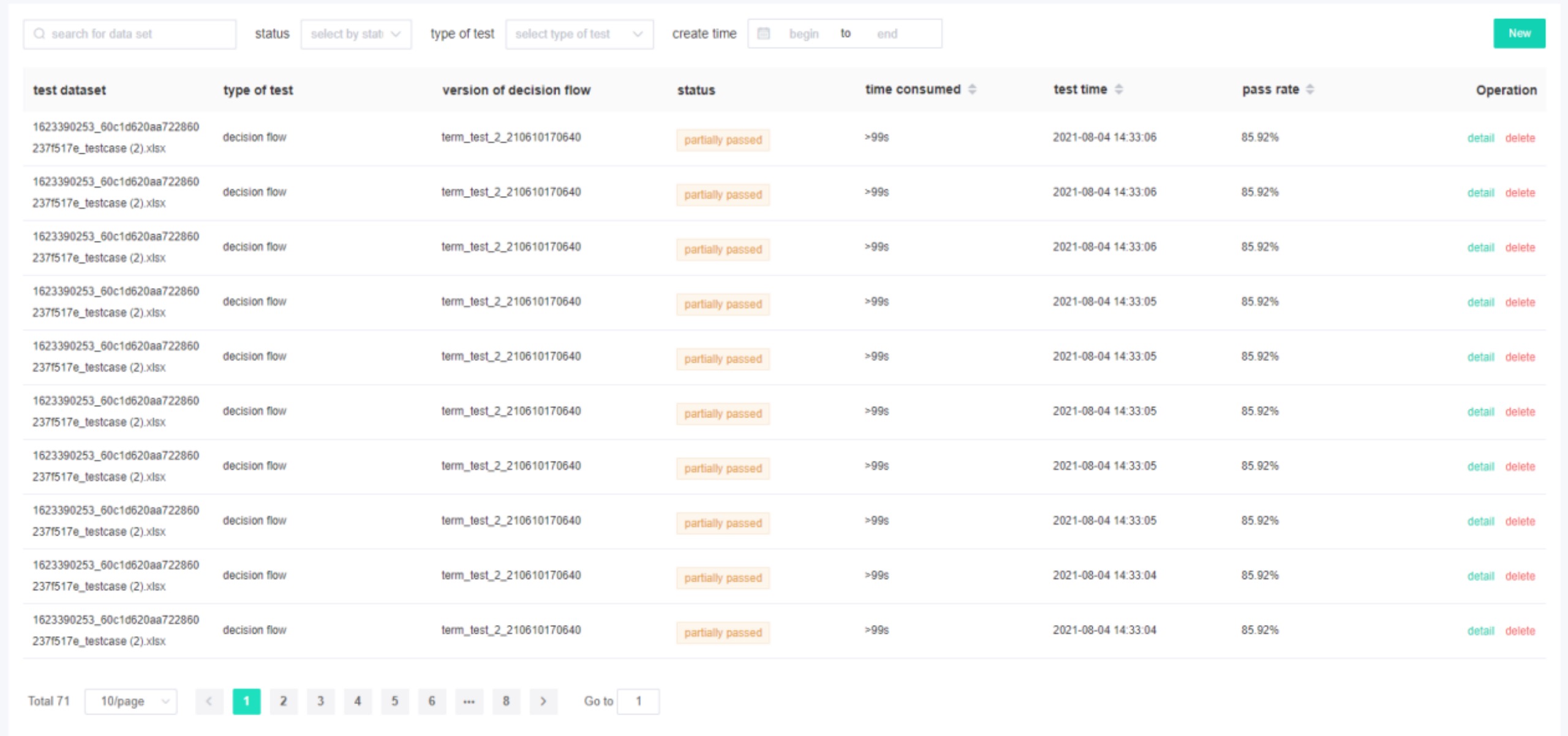}
    \caption{Statistics for testing records}
    \label{fig:galaxy}
\end{figure}

\subsubsection*{Action Box}
For better communicating with outer systems, we create a innovative way of responding to requests. For instance, an action of clinical decision support of diagnosing and making prescriptions can be applied by using it. After specifying the reasoning rules, the customized data structure of details in the prescription can be organized in it if it can be described in fixed rules. User can customize their responding structural data format in an easy way, we call this function Action Box. In which, two different types of responding format are included. User can use either synchronized or asynchronized responding format. In asynchronized one, URL for calling back needs to be specified. 

\begin{figure}[htp]
    \centering
    \includegraphics[width=12cm]{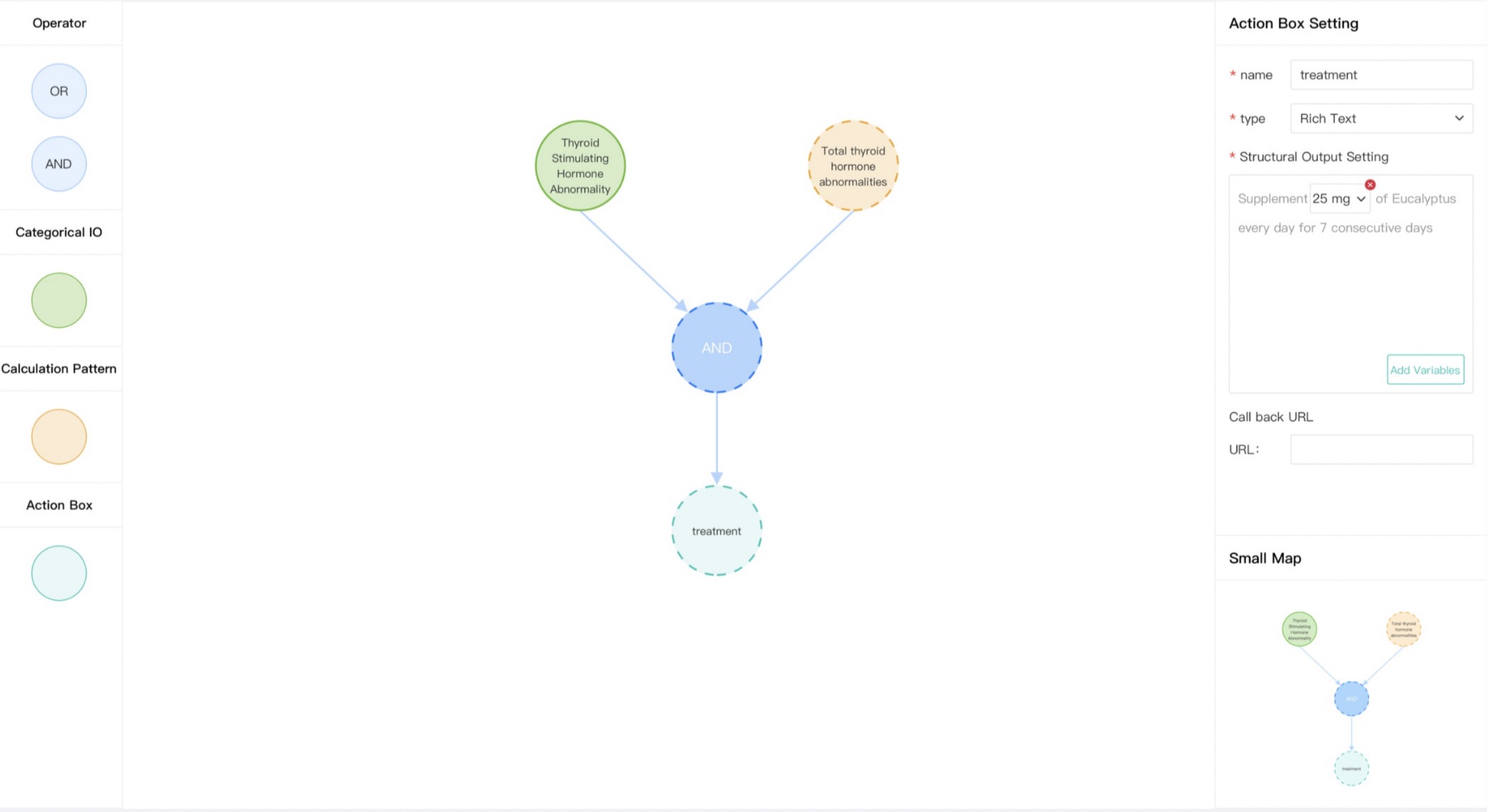}
    \caption{Action box for structural output}
    \label{fig:galaxy}
\end{figure}

\subsubsection*{Version Management}
Decision flow a basic unit for offering functions to outer systems. It implies that once the decision flow is open to public, it will be connected closely. Then slightly change may cause a lot problems. However, most of the the reasoning logic and medical knowledge are continuously being upgraded for delivering better medical service to patients. Therefore, publishing different versions for outer systems is needed. In our system, after the decision flow is built, it can be saved as a version. The testing process mentioned in the last sub chapter is based on a certain version. If one of the versions are tested and ready for publishing, it can be deployed. Other systems can invoke the published decision flow through HTTP protocol. Moreover, all of the versions published will be managed in a unified page, users can either terminate or activate them.

\begin{figure}[htp]
    \centering
    \includegraphics[width=12cm]{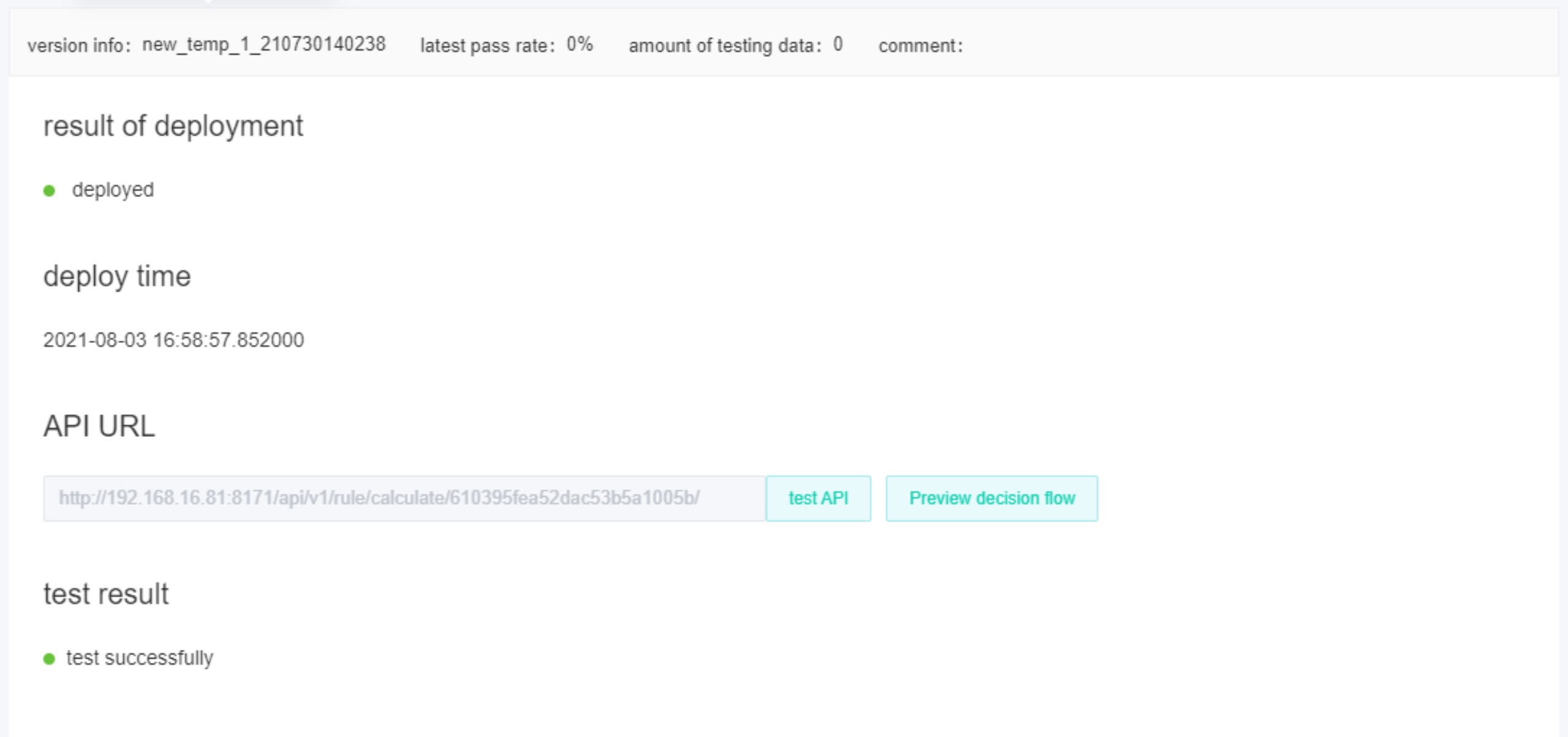}
    \caption{Publishing version}
    \label{fig:galaxy}
\end{figure}

\begin{figure}[htp]
    \centering
    \includegraphics[width=12cm]{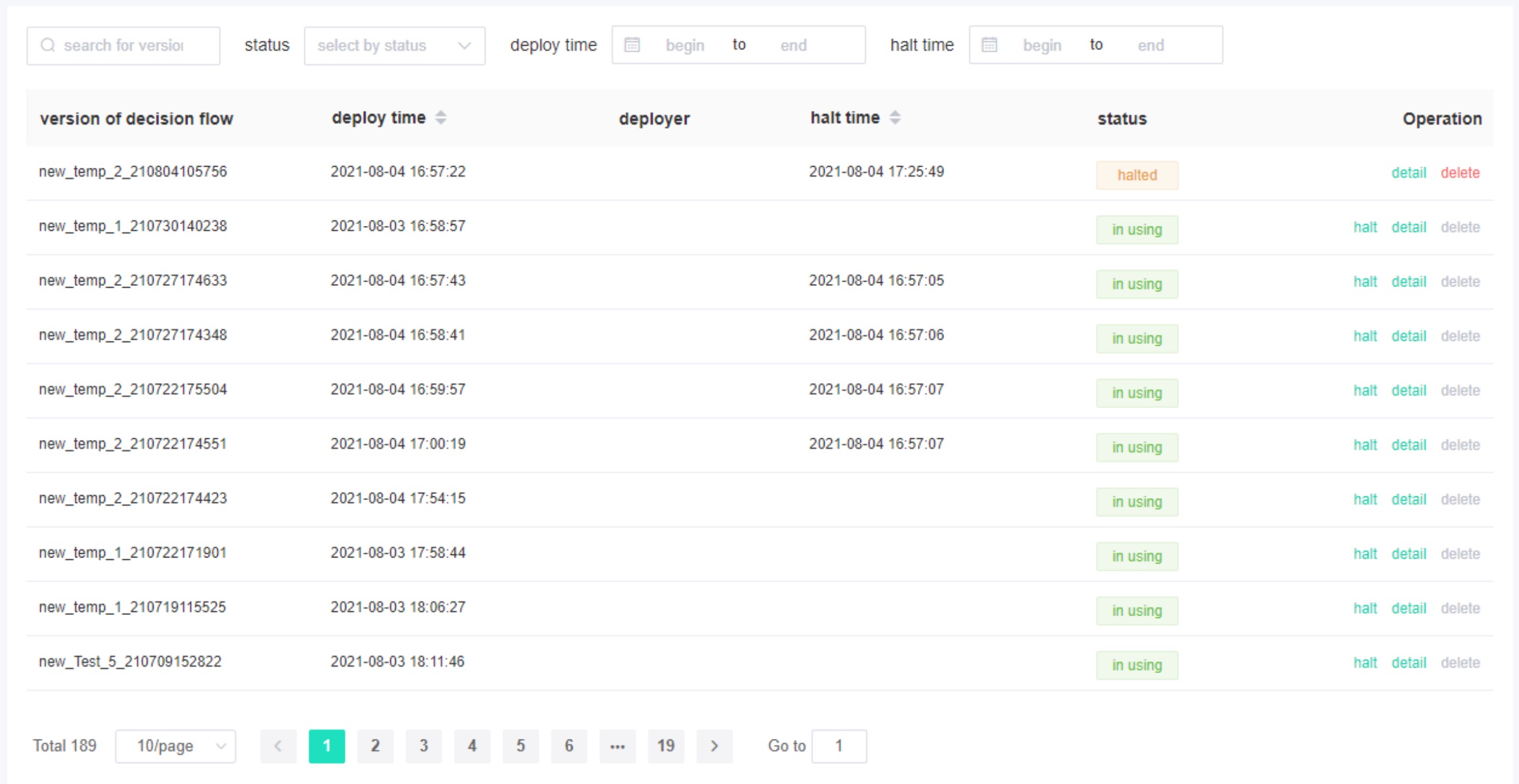}
    \caption{Management of published versions}
    \label{fig:galaxy}
\end{figure}

\subsection*{Architecture}
Knowledge Engine provides the ability of knowledge computation, which comes from the fundamental computational engine driven by logical reasoning rules. There are three major abilities in it, which are knowledge producing, knowledge computing and knowledge applying. With an intuitive knowledge inputting interface, user can extract information from guidelines or academic researches manually and organize them into knowledge in the form of reasoning rules. By using the fundamental powerful reasoning engine, static rules can be transformed into dynamic computational pathways. Knowledge is applied in practice by deploying to production environments, then combined as services for different practical scenarios, e.g. consultation systems or follow-up plans after treatment. The architecture diagram below shows the entire knowledge processing flow from knowledge producing to knowledge applying. 

\begin{figure}[htp]
    \centering
    \includegraphics[width=12cm]{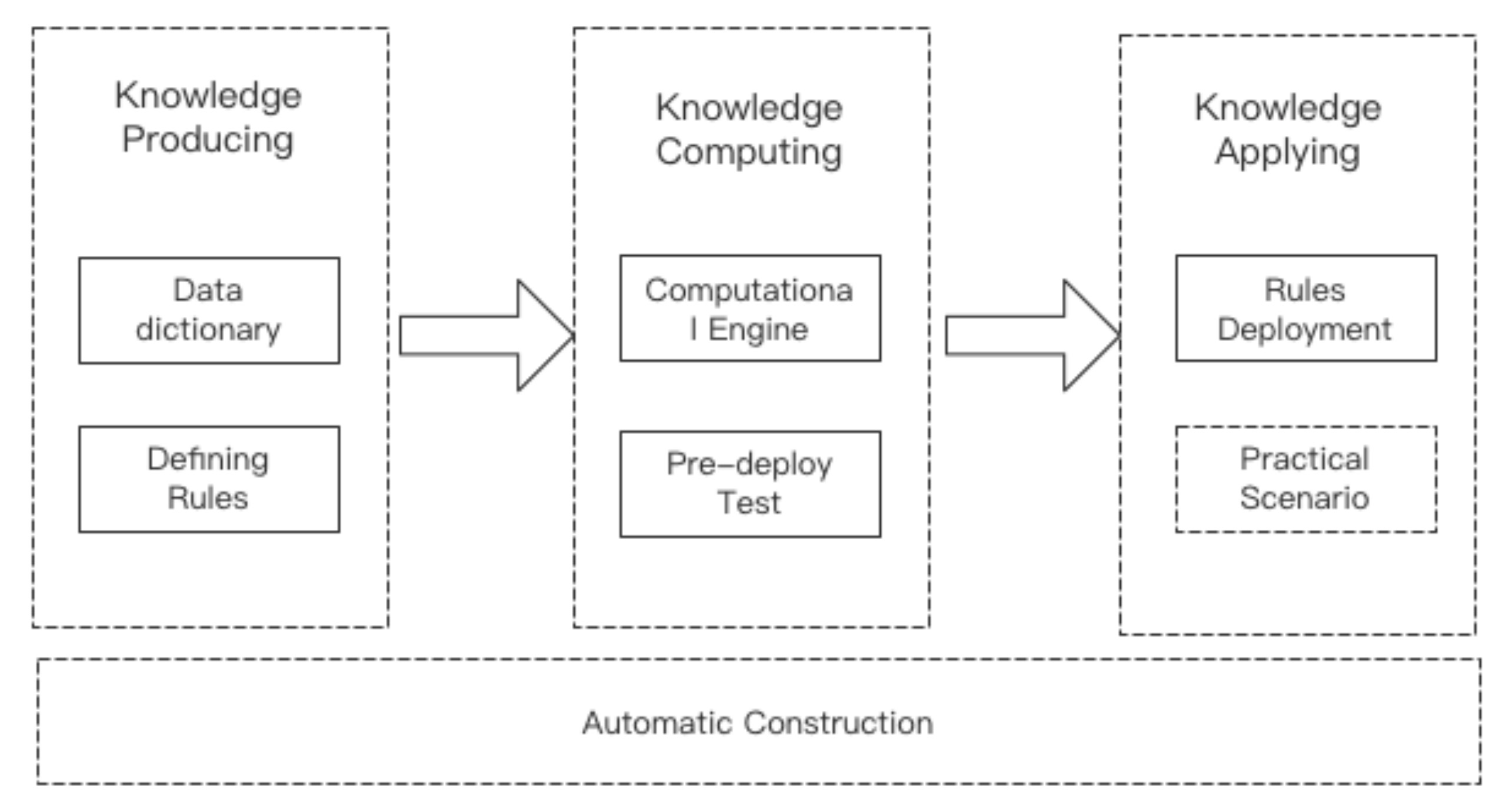}
    \caption{Knowledge Processing Flow}
    \label{fig:galaxy}
\end{figure}

Knowledge engine system is constructed in the form of micro services. Mainly, it includes the gateway services, front-end services, cache services, data persistent services, knowledge application management services and core reasoning engine services. Gateway services are in charge of providing communication and data exchanging abilities. Front-end services offer users an interface to interact with the system. Cache services are responsible for storing frequently used data in memories in order to gain high performance. Data persistent services are mainly for storing important data, including rules, logs etc, into reliable data base. Knowledge application management services are responsible for knowledge inputting, rules validating and standardizing. The core reasoning engine is used for supporting massive computational requests. The following diagram shows the structure mentioned above.

\begin{figure}[htp]
    \centering
    \includegraphics[width=12cm]{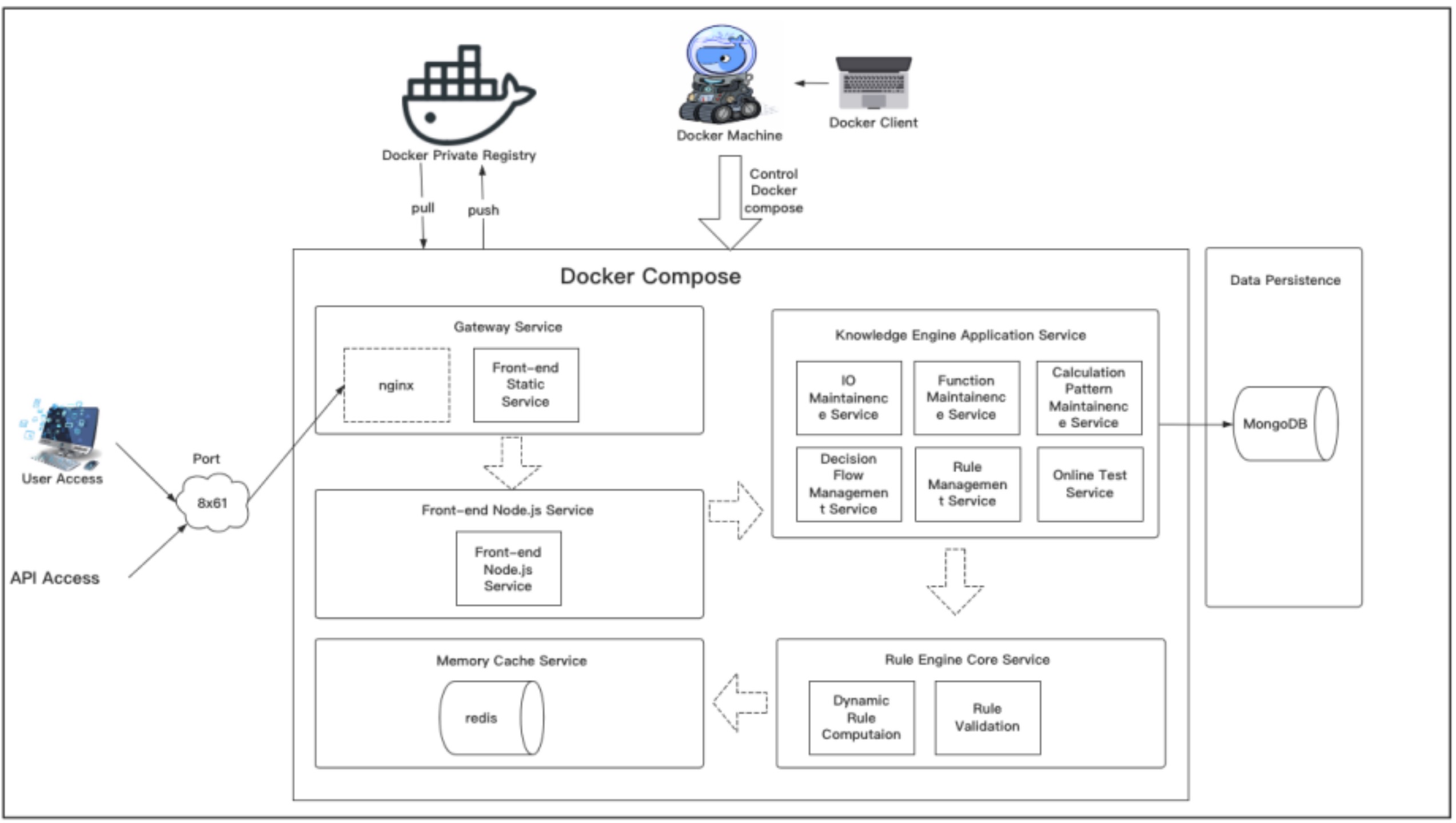}
    \caption{System Topology}
    \label{fig:galaxy}
\end{figure}

\subsubsection*{The Principle of Logical Reasoning}
Knowledge engine is a system developed based on the principle of general description of logic calculations, and replaces the traditional single, non-intuitive logic operation symbols with intuitive and highly interpretable logical interface languages. Including formal logic, numerical calculation and set operation.

It takes formal logic as the core. In formal logic, common logical operators are AND, OR, NOT. The design centered on formal logic enables knowledge engine to cover many medical knowledge inference scenarios. Use standard terminologies with logical operators and hypothetical reasoning models to draw conclusions, and conclusions are also expressed in medical terms. For example: A AND B -> C. In the above-mentioned conducting process, A, B, C are all expressed in medical terms. Besides, A and B are used as conditions to reach the conclusion C. In addition, testing function included in it follows De Morgan's law. Under the premise that the conclusion C is not established, it can be deduced back to show that one of the conditions A or B is not established or neither is established. In addition to formal logic, it also contains numerical calculations and set operations. Numerical calculation is a calculation method for terms with numerical values, and is implemented by the calculation pattern component mentioned above. Another type of operation that uses the calculation pattern component is set operations. It is mainly used to solve the problem of scale calculation in most medical scenarios. The problem is like: if the patient has headache, fever, sneezing, coughing, fatigue any three or more of the above will require further inspection.

\subsubsection*{The Fundamental Logic Computation Module}
The computing function of the knowledge engine is based on its underlying logic computing module. It adopts a decoupled design. The underlying logic computing module serves as an independent service module to provide underlying logic computing services for the knowledge engine and respond to specific computing requests. It supports basic formal logic, numerical and set operations. Plus, it supports multiple types of variables, including number, string, time stamp etc. 

There are two types of rules loading strategies, temporal and long-term. When the quick online testing or the batch testing is performed, rules will be loaded into memory temporally. Once the testing process is terminated, memory occupied will be released for saving space. If a decision flow is deployed to the public, rules built in it will be loaded into memory for long-term use in order to increase the efficiency rather than loading them repetitively. In addition, for further increasing the efficiency, rules will be pre-processed and optimized by flattening them horizontally in the form of linked table. Elements in the linked table are considered to be in different independent groups. Rules in the same group will be computed parallelly, but rules within different groups will be computed by respecting to the order of linked table, because there are intermediate values need to be outputted. One optional computing strategy can be set manually by users is storing the intermediate values in a session. It’s mainly used for connecting to HIS or other business systems because decision flows might have chance to be invoked repetitively, intermediate values can help saving computational resources. Moreover, intermediate value setting is on by default in the testing modules because back tracking is needed for showing the conducting path. 

\begin{figure}[htp]
    \centering
    \includegraphics[width=12cm]{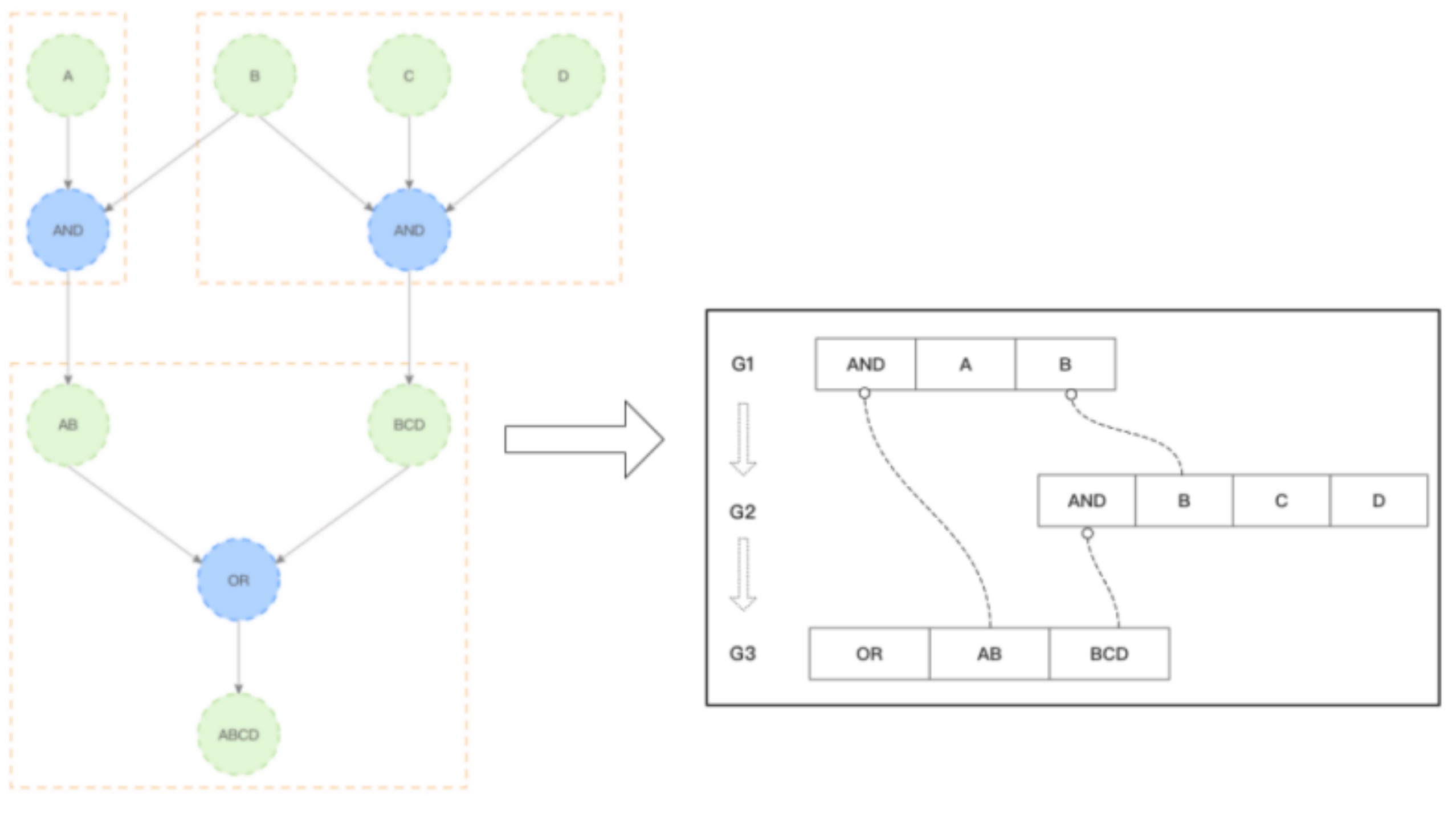}
    \caption{Transforming rules to data structure}
    \label{fig:galaxy}
\end{figure}

\subsubsection*{Design of Data Model}
The data model of knowledge engine is the abstract design for storing decision flows, decision blocks and specific rules data, which includes the following basic elements. IO is the particle of computing and it could be taken as variables in programming languages. It supports numerical, tag and set types. Operation is connector of IOs. It expresses the logical operations. It has two different types, AND and OR. Calculation pattern is a single computation unit used for numerical and set operations. The output of it must be a specified IO, therefore the result of calculation pattern could be fed into logical computation process directly. Rule is the smallest component which can be computed. It could be combined with IOs, operations and calculation patterns. The decision block can be seen as a box of rule which can be connected to form a complete decision flow. Decision flow is the smallest component which could be invoked by outer systems. The design of data model is shown as follows.

\begin{figure}[htp]
    \centering
    \includegraphics[width=12cm]{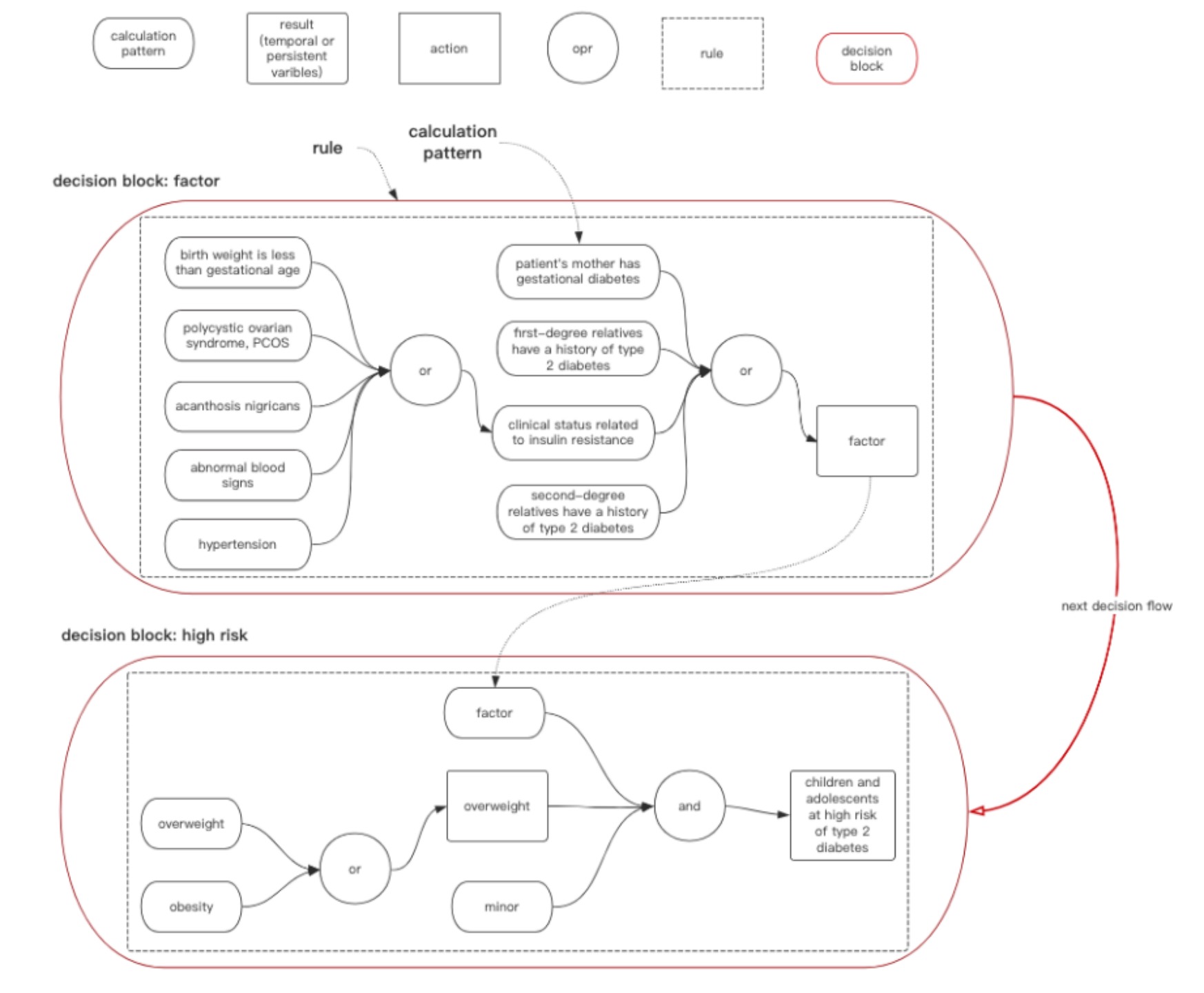}
    \caption{Data Model}
    \label{fig:galaxy}
\end{figure}

\subsubsection*{Design of Availability}
In order to meet the demand for availability, we adopt the following high-availability design. The knowledge engine system adopts modular design and they do not affect each other. This design pattern can prevent the failure of a single service from causing the entire system to be globally unavailable. In addition, the design of module separation allows each module to be dynamically and automatically expanded according to demand of expanding capacity of data processing. Services containerization and service orchestration are based on container service orchestration tools. Environment isolation and resource isolation between services facilitate rapid deployment, management and capacity expansion. The front end and back end services are separated, and they communicate with each other through a unified communication protocol format based on the HTTP protocol, which has high maintainability. The application service is separated from the logic computation. Data exchanged between them is storing in a middle cache to avoid the blocking problem caused by the imbalance of the upstream and downstream data flux. Web services and task scheduling are decoupled. The communication between them is carried out through queues to improve user experience.  Internal and external gateways are isolated for reasons of security.

\section*{Performance}
\label{sec:headings}
For evaluating the performance of knowledge engine, we have designed three experiments. They share the same settings of software and hardware. The deployment is as figure shown above. Nginx is as the load balancer which is responsible for accepting requests and responding. Services are deployed in multiple containers. We use Redis as the cache storing service. And we use Jmeter for simulating real user sending requests. We use four servers as the testing environment. The environmental settings of them is shown as follows.

\begin{table}[]
\centering
\begin{tabular}{|l|l|l|l|l|}
\hline
\textbf{\#} & \textbf{Deployments}                   & \textbf{CPU} & \textbf{Memory} & \textbf{OS}       \\ \hline
1           & Deploying Knowledge Engine with docker & 32           & 16G             & x86\_64 GNU/Linux \\ \hline
2           & MongoDB, Redis                         & 16           & 16G             & x86\_64 GNU/Linux \\ \hline
3           & Jmeter client                          & 16           & 16G             & x86\_64 GNU/Linux \\ \hline
4           & Prometheus monitoring system           & 32           & 16G             & x86\_64 GNU/Linux \\ \hline
\end{tabular}
\caption{Settings of hardware environment}
\end{table}

About the experimental parameter settings, we set it into 4 parts which are shown in the table below. Input IO stands for the original input variables, not including complex calculation patterns. Operators are logical operators, which are OR and AND. And the number of decision blocks will be shown as well.

\begin{table}[]
\centering
\begin{tabular}{|c|c|c|c|c|c|c|c|c|c|}
\hline
\multicolumn{10}{|c|}{\textbf{Settings of Experiments}}                                                                                       \\ \hline
Experiment         & \multicolumn{2}{c|}{\#Input IO} & \multicolumn{4}{c|}{\#Calculation Pattern} & \multicolumn{2}{c|}{\#Operator} & \#Block \\ \hline
\multirow{2}{*}{1} & \multicolumn{2}{c|}{100}        & \multicolumn{4}{c|}{0}                     & \multicolumn{2}{c|}{9}          & 1       \\ \cline{2-10} 
                   & negative * 0   & positive * 100 & /       & /       & /         & /          & OR * 6         & AND * 3        & /       \\ \hline
\multirow{2}{*}{2} & \multicolumn{2}{c|}{29}         & \multicolumn{4}{c|}{17}                    & \multicolumn{2}{c|}{27}         & 4       \\ \cline{2-10} 
                   & negative * 10  & positive * 19  & GT * 5  & TD * 4  & IntS * 3  & NumOp * 5  & OR * 12        & AND * 15       & /       \\ \hline
\multirow{2}{*}{3} & \multicolumn{2}{c|}{29}         & \multicolumn{4}{c|}{17}                    & \multicolumn{2}{c|}{27}         & 4       \\ \cline{2-10} 
                   & negative * 10  & positive * 19  & GT * 5  & TD * 4  & IntS * 3  & NumOp * 5  & OR * 12        & AND * 15       & /       \\ \hline
\end{tabular}
\caption{Settings of Experiments}
\end{table}

In experiment 1, we deploy 8 computational services and 8 gateway services for it. And we have tested it with 61590 samples in 3 minutes and 49 seconds. The average of responding time is 283.3 ms and the throughput per second is 269.17. During this experiment, the total use of CPU in No1 server is 26.38\% and the use of memory is 30\%. 

\begin{figure}[htp]
    \centering
    \includegraphics[width=12cm]{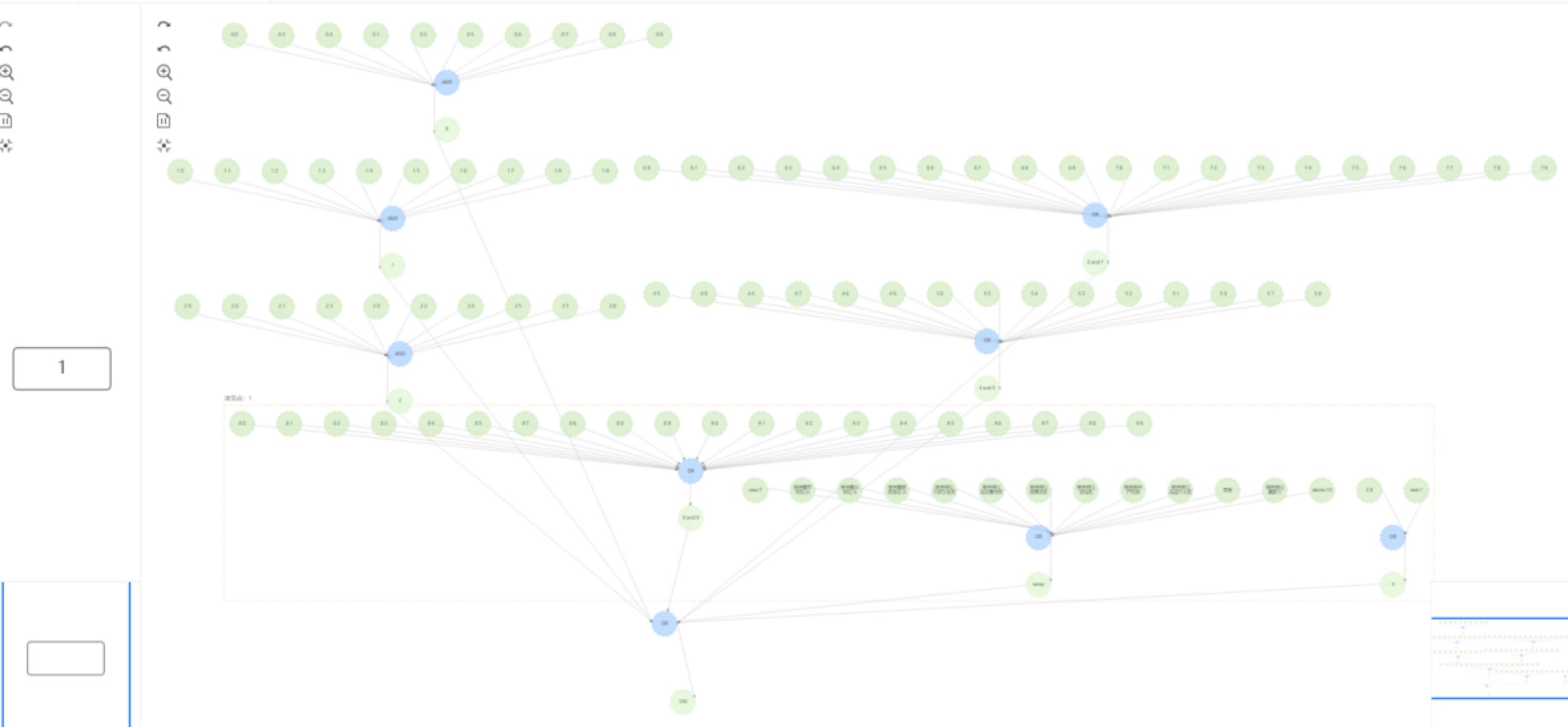}
    \caption{Reasoning rules in Experiment 1}
    \label{fig:galaxy}
\end{figure}

In experiment 2, We have tested it with 67694 samples in 3 minutes and 49 seconds. The average of responding time is 257.62 ms and the throughput per second is 295.84. During this experiment, the total use of CPU in No1 server is 26.15\% and the use of memory is 30\%. 

In the experiment 3, we set the decision flows the same as experiment 2. But in this experiment, we evaluate the performance after we have expand the number of servers to 2. We have tested it with 230200 samples in 6 minutes and 49 seconds. The average of responding time is 153.75 ms and the throughput per second is 563.19. During this experiment, the total use of CPU in the two servers are 28.55\% and 27.44\% and the use of memory are both 30\%. 

\begin{figure}[htp]
    \centering
    \includegraphics[width=12cm]{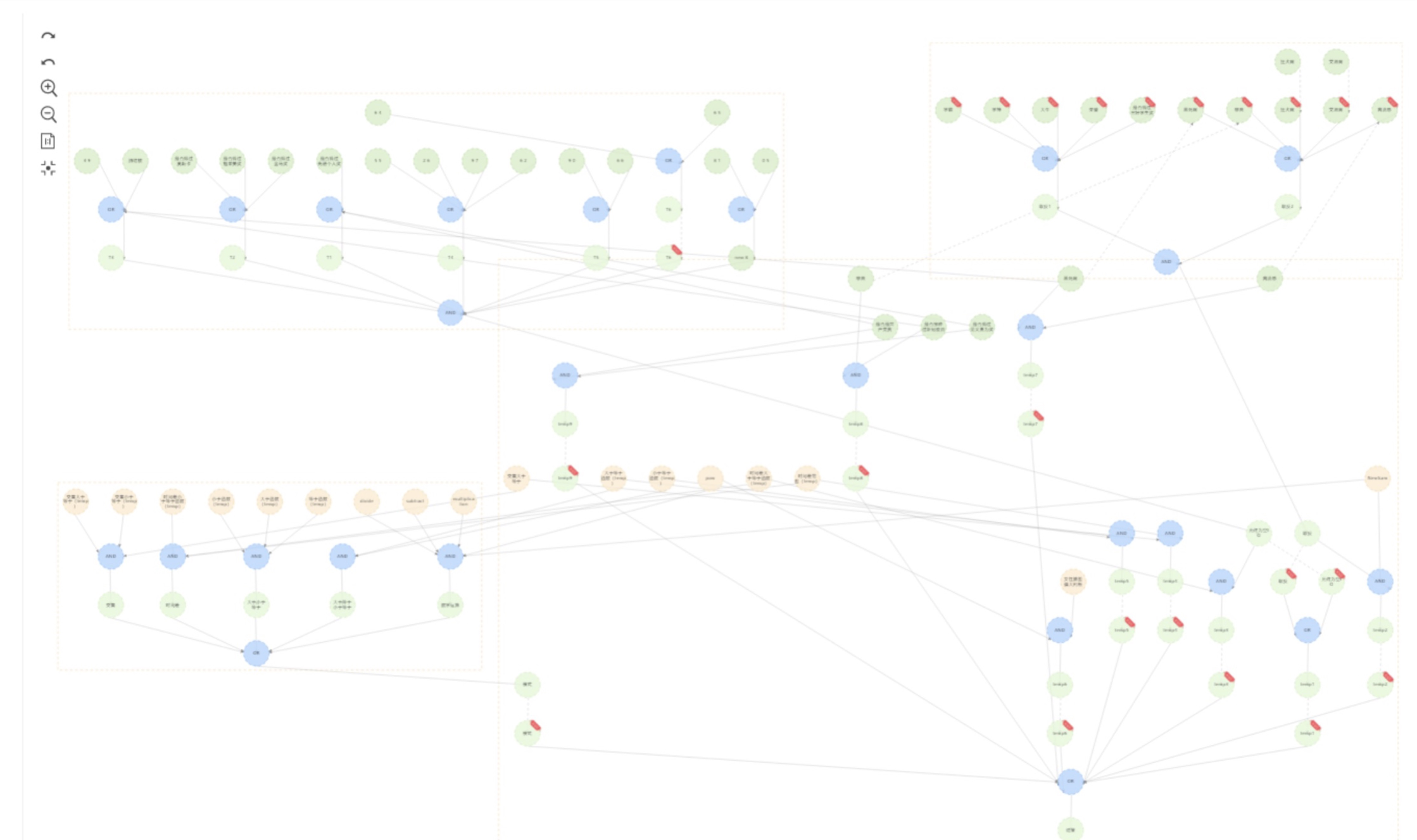}
    \caption{Reasoning rules in Experiment 2 and 3}
    \label{fig:galaxy}
\end{figure}

\begin{table}[]
\centering
\begin{tabular}{|c|c|c|c|c|c|c|}
\hline
\multicolumn{7}{|c|}{\textbf{Performance}}                                                                                                                  \\ \hline
\textbf{Experiment} & \textbf{\#Samples} & \textbf{Time} & \textbf{Avg. responding time(ms)} & \textbf{Throughput(TPS)} & \textbf{CPU}    & \textbf{Memory} \\ \hline
1                   & 61590              & 3'49''        & 283.21                            & 269.17                   & 26.38\%         & 30\%            \\ \hline
2                   & 67694              & 3'49''        & 257.62                            & 295.84                   & 26.15\%         & 30\%            \\ \hline
3                   & 230200             & 6'49''        & 153.75                            & 563.19                   & 28.55\%/27.44\% & 30\%            \\ \hline
\end{tabular}
\caption{Performance of each experiment}
\end{table}

The cases we design to take the experiment is more complex than the decision flows we have built in practice. The decision flow in practice has xxx IOs and xxx operators in it on average. We can see that the system in this specific settings are of high performance and is enough for regular scenarios. Moreover, users can improve the performance by expanding the number of servers. As it is depicted in the experiment 3, we gain approximately 90\% increasing of the throughput by adding one more server.

\section*{Conclusion}
In this paper, we present our innovative knowledge engine called Sinoledge. Firstly, we discuss the requirement of organizing thoughts, reasoning paths in the daily work of doctors, physicians and researchers. There are many potential scenarios where a powerful knowledge engine can be applied in, which are supporting clinical decision making, scheduling follow-up plans, conducting clinical trials etc. Secondly, some related work are shown, including clinical supports for different diseases, ways of connecting to outer systems which makes them more practical, architecture of the entire systems etc. Thirdly, we show the its major functionality and powerful architecture. It includes an intuitive user language for organizing thoughts and reasoning paths, an understandable testing mechanism, and a code-less way of managing decision flows and deployments. Lastly, a test for showing performance of our system is presented, in which we can see that our proposal is efficient and effective enough to be applied in practice.

\bibliographystyle{unsrt}  
\bibliography{references}

\end{document}